%% file: main.tex
\documentclass[10pt,twocolumn,letterpaper]{article}

\usepackage[pagenumbers]{cvpr} % To force page numbers, e.g. for an arXiv version

\usepackage{graphicx}
\usepackage{amsmath}
\usepackage{amssymb}
\usepackage{booktabs}

\DeclareMathOperator*{\argmin}{arg\,min}

\usepackage{mathtools}
\usepackage{bm}

\usepackage[pagebackref,breaklinks,colorlinks]{hyperref}

\usepackage[capitalize]{cleveref}
\crefname{section}{Sec.}{Secs.}
\Crefname{section}{Section}{Sections}
\Crefname{table}{Table}{Tables}
\crefname{table}{Tab.}{Tabs.}

\def\cvprPaperID{2} % *** Enter the CVPR Paper ID here
\def\confName{CVPR}
\def\confYear{2022}

\begin{document}

%%%%%%%%% TITLE - PLEASE UPDATE
\title{Privacy Leakage of Adversarial Training Models in Federated Learning Systems}

\author{Jingyang Zhang, Yiran Chen, Hai Li\\
Department of Electrical and Computer Engineering, Duke University\\
{\tt\small jingyang.zhang@duke.edu}
% For a paper whose authors are all at the same institution,
% omit the following lines up until the closing ``}''.
% Additional authors and addresses can be added with ``\and'',
% just like the second author.
% To save space, use either the email address or home page, not both
%\and
%Second Author\\
%Institution2\\
%First line of institution2 address\\
%{\tt\small secondauthor@i2.org}
}
\maketitle

%%%%%%%%% ABSTRACT
\begin{abstract}
   Adversarial Training (AT) is crucial for obtaining deep neural networks that are robust to adversarial attacks, yet recent works found that it could also make models more vulnerable to privacy attacks.
   In this work, we further reveal this unsettling property of AT by designing a novel privacy attack that is practically applicable to the privacy-sensitive Federated Learning (FL) systems.
   Using our method, the attacker can exploit AT models in the FL system to accurately reconstruct users' private training images even when the training batch size is large.
   Code is available at \url{https://github.com/zjysteven/PrivayAttack_AT_FL}.
\end{abstract}
\vspace{-5mm}

%%%%%%%%% BODY TEXT
\input{sections/1_introduction}

\input{sections/2_related_work}

\input{sections/3_setting}
\input{sections/4_methodology}

\input{sections/5_experiments}
\input{sections/6_conclusion}

\clearpage
%%%%%%%%% REFERENCES
{\small
\bibliographystyle{ieee_fullname}
\bibliography{egbib}
}

\end{document}

%% file: sections/1_introduction.tex
\section{Introduction}

Deep Neural Networks (DNNs) suffer from the notorious adversarial perturbations \cite{adv_example}: These imperceptible noises could easily fool DNNs to yield wildly wrong and malicious decisions (\eg, misclassification with high confidence).
\textit{Adversarial Training} (AT) \cite{Advt_madry} has been one of the most effective techniques that mitigate such vulnerability, which withstands adaptive attacks \cite{tramer2020adaptive} and leads to the highest empirical adversarial robustness to date \cite{croce2020robustbench}.
It is without doubt that AT is crucial for building robust intelligent systems.

Despite the desired robustness, recent works \cite{song2019privacy,mejia2019robust} revealed an unexpected and alarming property of AT: It can make models more likely to leak the data privacy than the ones that undergoes vanilla training (\ie, using the original clean images instead of online-generated adversarial examples).
Specifically, Song \etal \cite{song2019privacy} found that AT makes it easier for \textit{membership inference attack} \cite{shokri2017membership} to succeed, which aims to identify whether a certain sample was used during the training.
In another work \cite{mejia2019robust}, it is shown that \textit{model inversion attack} \cite{fredrikson2015model} is tractable on AT models, with which the attacker can generate images that visually resemble the actual training samples.
%exhibit visual resemblance to the actual samples from the training dataset. 
In all, both papers indicated that AT models exhibit a robustness-privacy trade-off.

%A later work \cite{helland2020human} investigated the ``human-recognizability phenomenon'' of AdvT models and advocated for 

\begin{figure}[!t]
  \centering
  \includegraphics[width=.9\linewidth]{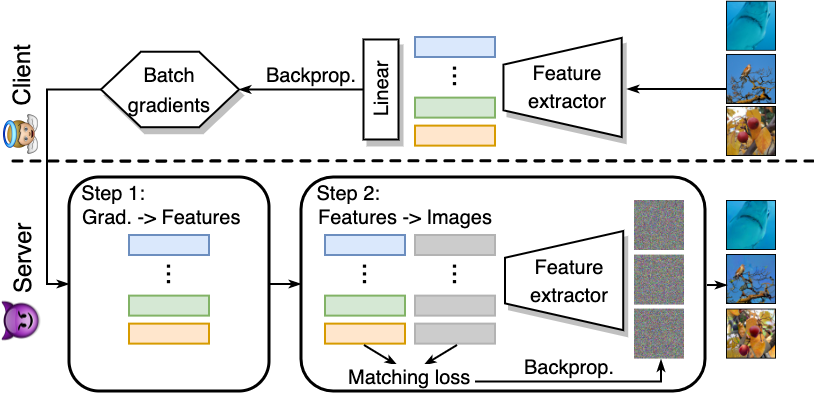}
  %\vspace{-1mm}
  \caption{An overview of the proposed privacy attack against FL systems to ``steal'' the client's training images. We formulate a novel two-step procedure: 1) Feature restoration from gradients (\cref{sec:4.2}) and 2) Image reconstruction from features (\cref{sec:4.3}). Zoom in and also refer \cref{fig:demo} to see image reconstruction quality.}
  \label{fig:overview}
  \vspace{-5mm}
\end{figure}

In this work, we pursue this line of inquiry and further demonstrate such trade-off by presenting a novel privacy attack that exploits AT models.
Our attack is practically applicable to \textit{Federated Learning} (FL) systems \cite{FL}. 
The goal is to compromise the privacy of FL clients by reconstructing their own training images (which may contain private information) based on the weight gradients that are communicated between the clients and the central server.

Different from previous approaches that directly invert gradients to reconstruct images \cite{zhu2020deep,geiping_inverting_grad,yin2021see}, our attack methodology forms a unique two-step procedure, where we first restore the features (\ie, the outputs of the penultimate layer) from the gradients, and then reconstruct the inputs using the recovered features as supervision (see \cref{fig:overview} for illustration).
The key insight behind our method is that while the gradients of the samples within a batch are fused together and prohibit accurate reconstruction of each individual input, the features are less coupled and (once restored) can provide more precise information for the reconstruction than the batch-averaged gradients.
Upon the general framework, the proposed attack leverages AT models which enable the success execution of both attack steps.
%which we will discuss later in detail.

We conduct extensive experiments using high-resolution images from ImageNet \cite{deng2009imagenet} to demonstrate how our privacy attack can work with AT models to compromise the privacy of FL clients.
Specifically, we are able to accurately reconstruct the clients' training images at a large batch size (\eg, 256), something that is not achieved by previous methods with vanilla models \cite{zhu2020deep,geiping_inverting_grad,yin2021see}.
The results suggest that AT poses a risk to the expected-to-be-secure FL systems.

%% file: sections/2_related_work.tex
\section{Related Work}

\noindent \textbf{Privacy risks of AT models.}
As discussed earlier, a few prior works \cite{song2019privacy,mejia2019robust} also identified that AT could make models more vulnerable to privacy attacks.
In comparison, our work focuses on a considerably more strict and challenging criterion of privacy leakage: We consider exact reconstruction of the training images owned by the victims, a scenario which we believe poses more severe risks than the attacks that synthesize human-recognizable images \cite{mejia2019robust} or infer the membership of data samples \cite{song2019privacy}.
Besides, we put ourselves in the realistic, privacy-sensitive FL setting, to which our attack is practically applicable.
%The work of \cite{helland2020human} advocated for more research to better understand how privacy attacks are possible against AT models.

\noindent \textbf{Privacy attacks against FL systems.}
There has been a line of research trying to expose the privacy vulnerability of the FL systems.
Zhu \etal \cite{zhu2020deep} developed the seminal DLG attack which reconstructs input images by matching the gradients computed using the reconstructed ones and the real ones.
Geiping \etal \cite{geiping_inverting_grad} made improvements by using a better matching loss function and optimizer.
They achieved near-perfect recovery quality when the gradients are computed using a single image (\ie, the batch size is 1).
More recently, Yin \etal \cite{yin2021see} further incorporated advanced image priors as regularization to enhance the recover quality.

All these works, however, can only unveil clear visual information under a small batch size (\eg, when the batch includes only a single or a few images), while a large batch size (\eg, 128 or 256) is often required for effective and efficient training of DNNs \cite{he2016deep,KeskarMNST17}.
As a result, whether the privacy can really be compromised in realistic settings (\ie, with large batch size) remains unclear.
In this paper, we show that the privacy of FL clients is indeed at risk if they are performing AT, as the attacker can leverage our attack and exploit AT models to achieve accurate reconstruction of users' images even at a batch size of 256.

%it is recognized that the recovered images lack many details when the batch size is large (see {\color{blue} Fig. X}), and whether the privacy is really compromised in this more realistic case (compared with the single image case) is arguable.
%In this paper, we show that accurate image recovery in large batch is possible with AT models.

\noindent \textbf{Inverting features to reconstruct images.}
Inverting the features has been studied as a way to understand/interpret what DNN models learn \cite{mahendran2015understanding,engstrom2019adversarial}.
In particular, Engstrom \etal \cite{engstrom2019adversarial} discovered that AT models are more ``invertible'' than vanilla models, as the reconstructed images from AT models' features look much more plausible.
However, their focus is on the interpretability perspective.
In our work, we take the security perspective and demonstrate how the good ``invertibility'' of AT models can unexpectedly/undesirably enable a privacy attack.

%\textbf{Adversarial training and its properties.}
%AT; perceptually-aligned gradients; privacy issues

%\textbf{Visual privacy leakage in FL systems.}

%% file: sections/3_setting.tex
\section{Problem Setting}

We first formulate the problem setting and define the threat model before delving into our proposed attack. 
In a FL system, in each step the client will receive a copy of the global model from the central server and perform local training.
Here, following prior works \cite{zhu2020deep,geiping_inverting_grad,yin2021see}, we focus on the case of single-step local training.
Concretely, considering an image classification task, the gradients of loss w.r.t. the model parameters are computed as follows during the local training:
\begin{equation}
    \Delta_\theta \coloneqq \frac{1}{N}\sum_{i=1}^{N}\nabla_{\theta}\mathcal{L}(f_{\theta}(\bm{x}_i),y_i),
\label{eq:batch_grad}
\end{equation}
where $f_\theta(\cdot)$ is the mapping function of the DNN model parameterized by $\theta$, $N$ is the batch size, $\bm{x}_i$ is $i$-th input image of the batch, $y_i$ is the corresponding groundtruth label, $\mathcal{L}(\cdot, \cdot)$ computes the cross-entropy loss, and $\Delta_\theta$ is a simplified notion for the batch-averaged gradients.
Note, if the client is performing AT, the input $\bm{x}_i$ will be the adversarial example instead of the original clean sample.

After the local training, the server will collect the gradients $\Delta_\theta$ from the clients and aggregate the updates into the current global model $f_\theta(\cdot)$.
Here, we assume that the server is malicious or has been compromised by the attacker.
Therefore, the attacker can access the model $f_\theta(\cdot)$ and the gradients $\Delta_\theta$.
Similar to \cite{geiping_inverting_grad}, we do \textit{not} pose any further assumptions beyond this point which may allow the attacker to better exploit the vulnerability, \eg, we do not assume that the attacker can modify the model architecture or send fake, malicious global parameters to the clients.
The attack goal is then to reconstruct the input images $\bm{x}_i$ that are used during the last local training step, given the available information $f_\theta(\cdot)$ and $\Delta_\theta$.

%% file: sections/4_methodology.tex
\section{Methodology}
\label{sec:method}

\subsection{Overview}
To better motivate and describe our proposed methodology, we first give a brief overview of existing methods \cite{zhu2020deep,geiping_inverting_grad,yin2021see} and discuss their shortcomings.
Essentially, previous works all solve the following optimization problem:
\begin{equation}
    \argmin_{\bm{z}}\mathcal{L}_{\text{grad}}\left(\frac{1}{N}\sum_{i=1}^{N}\nabla_{\theta}\mathcal{L}(f_{\theta}(\bm{z}_i),\hat{y}_i), \Delta_\theta \right)+\mathcal{R}(\bm{z}).
\label{eq:prior_method}
\end{equation}
Specifically, the optimization variable here is a randomly-initialized batch of inputs $\bm{z}$.
The first term in \cref{eq:prior_method} enforces the gradients of loss computed using $\bm{z}$ to match the ground-truth gradients $\Delta_\theta$ according to a certain distance metric $\mathcal{L}_\text{grad}(\cdot,\cdot)$ (\eg, cosine distance is used in \cite{geiping_inverting_grad} and $\ell_2$ distance is used in \cite{zhu2020deep} and \cite{yin2021see}).
Note, an estimation of the ground-truth labels $\hat{y}_i$ is used here to compute the loss \cite{zhao2020idlg,geiping_inverting_grad,yin2021see}.
The underlying idea of the first term is that the synthesized images $\bm{z}$ will resemble the original inputs if the gradients computed on them are similar.
Meanwhile, to encourage the generation of realistic and natural images, the second term in \cref{eq:prior_method} incorporates image priors $\mathcal{R}(\cdot)$ as regularization (\eg, the Total Variation loss \cite{mahendran2015understanding} and the Group Consistency loss \cite{yin2021see}).

While the idea of \cref{eq:prior_method} is natural and straightforward, one can identify an obvious limitation.
Note from \cref{eq:batch_grad} that the gradients contributed by each individual sample, \ie, $\nabla_{\theta}\mathcal{L}(f_{\theta}(x_i),y_i)$, are fused together in the batch-averaged gradients $\Delta_\theta$.
Therefore, $\Delta_\theta$ does not provide precise supervision for each individual image especially under a large batch size, making it much difficult to achieve accurate reconstruction (in a similar sense to how exactly knowing the value of $a$ and $b$ is hard given the value of $a+b$).
As a result, \cref{eq:prior_method} only works well when the batch size is extremely small (\eg, when the batch has only a single or a few samples).
When the batch size is large, the restored images lack clear visual details and do not resemble the real images anymore \cite{geiping_inverting_grad,yin2021see}.

To mitigate this problem, we propose a unique two-step attack procedure.
In the first step, we decouple the fused information of each individual image by restoring their features from the gradients which are less coupled.
Here, we define the feature vector of a sample as the output of the penultimate layer when that sample is passed through the DNN model.
Once we restore the feature vectors, each corresponding to one input image, we can use them as more accurate supervision to reconstruct the images.
Next, we will explain each of the two steps in detail and discuss how AT models kick in and benefit the attack.

\subsection{Feature restoration from gradients}
\label{sec:4.2}

We first introduce some notions to facilitate the description.
A DNN model $f(\cdot)$ deployed for image classification tasks can always be decomposed as $f(\cdot)=h(r(\cdot))$, where $r(\cdot)$ typically comprises convolutional layers and extracts high-level representations/features of the inputs, and $h(\cdot)$ is the linear layer that performs classification on top of the features and outputs activation scores for each class (\ie, logits).
For simplicity, we denote the feature of an input $\bm{x}$ as $\bm{r}$, \ie, $\bm{r}\coloneqq r(\bm{x})$.
Note that $\bm{r}\in\mathbb{R}^{D}$ is a $D$-dimensional column vector.

Our key insight here is that the features can be restored from the gradients w.r.t. the linear layer's weights, which are directly accessible to the attacker.
To see this, let us first delve into the computation of the weight gradients of $h(\cdot)$.
Denote the linear layer's weight matrix as $\bm{W}\in\mathbb{R}^{D\times K}$, wherein the $k$-th column $\bm{W}_k\in\mathbb{R}^{D}$ is a weight vector corresponding to the $k$-th class, and $K$ is the total number of classes.
Then, the class activations/logits are $\bm{a}=\bm{W}^\top\bm{r}$, and the class probabilities are $\bm{p}=\text{softmax}(\bm{a})$, where the probability on $k$-th class $\bm{p}_k$ is $\frac{e^{\bm{a}_k}}{\sum_{j=1}^K e^{\bm{a}_j}}$. 
Here, $\bm{a}_k=\bm{W}_{:,k}^\top\bm{r}$ is the $k$-th element of the vector $\bm{a}$, with $\bm{W}_{:,k}$ being the $k$-th column of $\bm{W}$. 
Finally, the cross-entropy loss computed on an input pair $(\bm{x},y)$ is $l=-\log\bm{p}_{y}$.
Accordingly, one can derive the gradients of loss w.r.t. each weight vector $\bm{W}_{:,k}$ in the linear layer:
\begin{align}
    \frac{\partial l}{\partial\bm{W}_{:,k}}&=\frac{\partial l}{\partial\bm{p}_y}\cdot\sum_{j=1}^{K}\frac{\partial\bm{p}_y}{\partial\bm{a}_j}\cdot\frac{\partial\bm{a}_j}{\bm{W}_{:,k}}\nonumber\\
    &=\begin{cases}
      -(1-\bm{p}_y)\bm{r}, & k=y\\
      ~~~~~~~~~~~~~\bm{p}_k\bm{r}, & k\neq y
    \end{cases}.
\label{eq:fc_grad}
\end{align}

Based on the derivation in \cref{eq:fc_grad}, we now discuss how one can restore the features from the batch-averaged gradients w.r.t. the linear layer's weights.
We consider a batch of $N$ inputs $\bm{x}_i$, each with a corresponding feature vector $\bm{r}_i$.
Note, following \cite{geiping_inverting_grad,yin2021see}, here we assume that there are no samples sharing the same label within the batch.
This assumption generally holds for the randomly constructed batch when the size $N$ is (much) smaller than the number of classes $K$ (\eg, on ImageNet $K=1000$ while $N$ typically takes 256 \cite{he2016deep}).
Then, suppose that we want to restore the feature vector $\bm{r}_i$ whose corresponding label is $y_i$, we can directly take the gradients w.r.t. the weight vector $\bm{W}_{:,y_i}$:
\begin{align}
    \hat{\bm{r}}_i &\coloneqq \Delta_{\bm{W}_{:,y_i}}=
    \frac{1}{N}\sum_{j=1}^{N}\frac{\partial l_j}{\partial \bm{W}_{:,y_i}}\nonumber\\
    &\propto -(1-\bm{p}_{i,y_i})\bm{r}_i+\sum_{j=1,j\neq i}^N\bm{p}_{j,y_i}\bm{r}_j\nonumber\\
    &\approx -(1-\bm{p}_{i,y_i})\bm{r}_i.
\label{eq:feat_restore}
\end{align}
Here, we denote the restored feature vector as $\hat{\bm{r}}_i$. 
$l_j$ is the loss computed on the $j$-th image, and $\bm{p}_{i,y_i}$ represents the probability of the $i$-th sample belonging to class $y_i$.

\begin{figure}[!b]
  \centering
  \includegraphics[width=.85\linewidth]{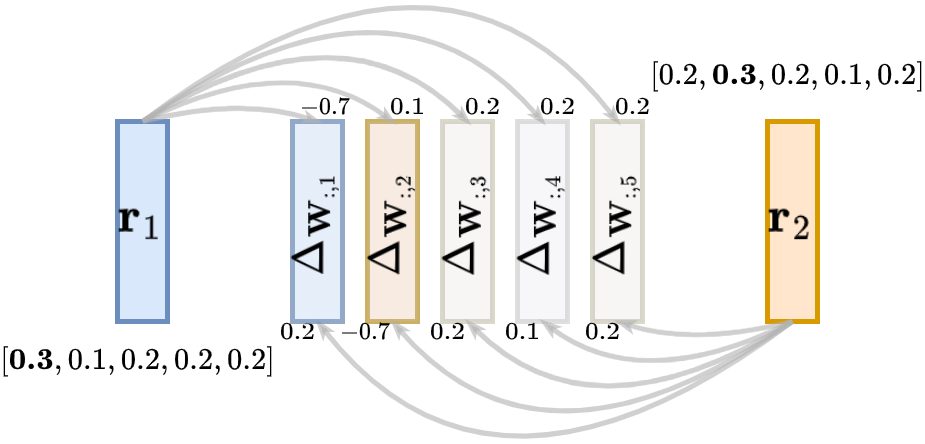}
  %\vspace{-1mm}
  \caption{A toy example showcasing how the features can be restored from the linear layer's weight gradients on an AT model. Here we suppose the batch size $N=2$ and the number of classes $K=5$. Without loss of generality, assume $\bm{r}_1$ and $\bm{r}_2$ have label $1$ and $2$, respectively. The numbers in the brackets are the softmax probabilities. Adversarial examples result in small $\bm{p}_{1,1}$ and $\bm{p}_{2,2}$.
  According to \cref{eq:feat_restore}, $\Delta_{\bm{W}_{:,1}}\propto-(1-0.3)\bm{r}_1+0.2\bm{r}_2$. The attacker can take $\Delta_{\bm{W}_{:,1}}$ as the restored feature $\hat{\bm{r}}_1$.}
  \label{fig:feat_restore}
  %\vspace{-6.5mm}
\end{figure}

Essentially, \cref{eq:feat_restore} indicates that the restored feature $\hat{\bm{r}}_i$ is approximately proportional to the true feature $\bm{r}_i$.
The approximation in the last row of \cref{eq:feat_restore} holds because the probabilities assigned to the wrong class $\bm{p}_{j,y_i}$ (recall that $y_j\neq y_i$ when $j\neq i$) are becoming smaller and smaller as the training proceeds, therefore they may contribute much less to the summation than $-(1-\bm{p}_{i,y_i})\bm{r}_i$, especially if $\bm{p}_{i,y_i}$ is not too large.
In fact, this is exactly the case for an AT model.
During each iteration, AT uses adversarial examples as inputs, which are generated by maximizing the cross-entropy loss.
Equivalently, adversarial examples minimize the probability assigned to the ground-truth class, \ie, $\bm{p}_{i,y_i}$.
As a result, AT models can lead to a more tight approximation of \cref{eq:feat_restore} and better restoration of the features.
We present a toy example of the feature restoration process in \cref{fig:feat_restore}.
Note, although in the case of AT the attacker is actually reconstructing the adversarial example, the privacy can still be compromised because the adversarial image is visually very similar to the clean image since their distance is bounded by a small value.

%{\color{blue} Later we will show that compared with vanilla models, AT models naturally leads to smaller $\bm{p}_{i,y_i}$ due to the use of adversarial examples, thus resulting in a more tight approximation and better restoration of the features.}

We finally discuss how to recover the batch labels $y_i$, which is necessary for knowing which columns of the gradient matrix of $\bm{W}$ should be taken for feature restoration.
Here we adopt the common practice of existing works \cite{zhao2020idlg,geiping_inverting_grad,yin2021see}, which is to look at the sign of the gradient elements of $\bm{W}$.
For the ease of illustration, without loss of generality, assume that the batch of $N$ samples are with the labels $1,2,...,N$, \ie, the $k$-th sample is from the $k$-th class.
Then according to \cref{eq:feat_restore}, the gradients w.r.t. the $k$-th column of $\bm{W}$ are:

\vspace{-5mm}
\begin{align}
    &\frac{1}{N}\sum_{j=1}^{N}\frac{\partial l_j}{\partial \bm{W}_{:,k}}\propto\nonumber\\
    &~~~~~~\begin{cases}
        -(1-\bm{p}_{k,k})\bm{r}_k+\sum_{j=1,j\neq k}^N\bm{p}_{j,k}\bm{r}_j, & ~1\leq k\leq N\\
        \sum_{j=1}^N\bm{p}_{j,k}\bm{r}_j, & N<k\leq K
    \end{cases}.
\label{eq:sign}
\end{align}
Note that modern DNN architectures (\eg, ResNet \cite{he2016deep} and VGG \cite{simonyan2014very}) typically use ReLU as the activation function, which makes the elements of the feature vector $\bm{r}$ always non-negative.
With this in mind, the takeaway of \cref{eq:sign} is that in the gradient matrix of $\bm{W}$, only the columns that correspond to the appeared labels may have negative elements, while the other columns will have all positive values.
Therefore, we can determine the batch labels by checking the sign of the smallest value of each column.
If the smallest value of $k$-th column is positive, then $k$ is very unlikely to be one of the batch labels.
In practice, one can pick the top-$N$ columns that have the smallest elements, and take their indices as the batch labels.

\subsection{Image reconstruction from features}
\label{sec:4.3}

Once we obtain the restored feature $\hat{\bm{r}}_i$, the attacker can reconstruct each input image by solving the following optimization objective:
\begin{equation}
    \argmin_{\bm{z}_i}\mathcal{L}_{\text{feat}}\left( r_\theta(\bm{z}_i),\hat{\bm{r}}_i \right)+\mathcal{R}(\bm{z}_i).
\label{eq:feat_2_image}
\end{equation}
The idea here is to optimize the input $\bm{z}_i$ such that its feature $r_\theta(\bm{z}_i)$ is close to the restored feature $\hat{\bm{r}}_i$ (which in turn is similar to the ground-truth feature $\bm{r}_i$) according to a certain metric $\mathcal{L}_{\text{feat}}(\cdot,\cdot)$.
Here we opt to use the scale-invariant cosine distance as $\mathcal{L}_{\text{feat}}$ since $\hat{\bm{r}}_i$ approximates $\bm{r}_i$ by a factor of $-(1-\bm{p}_{i,y_i})$ (\cref{eq:feat_restore}).
Again, image prior $\mathcal{R(\cdot)}$ is incorporated to improve the fidelity of the generated images.
In this work, we apply the Total Variation \cite{mahendran2015understanding} as the regularization, while using more sophisticated/complex image priors is likely to further enhance the reconstruction quality.

Inverting features to reconstruct input images is more feasible on AT models than on vanilla models.
This observation is first made in the work of \cite{engstrom2019adversarial}, where it is found that inverting the features of AT models can yield meaningful images that share a great amount of semantic similarity with the real inputs, while doing so on vanilla non-robust models only lead to meaningless noisy patterns.
Back then, the researchers focused on the interpretability aspect of such phenomenon without discussing its security implications.
In this work, we show how AT models' good ``invertibility'' can actually be exploited to conduct privacy attack.

%% file: sections/5_experiments.tex
\section{Experiments}

\subsection{Setup}

\noindent \textbf{Dataset.}
We target the reconstruction of $224\times 224$ high-resolution images from ImageNet \cite{deng2009imagenet}, which is more challenging than restoring low-resolution images (\eg, $32\times 32$ CIFAR images) \cite{geiping_inverting_grad}.

\noindent \textbf{Models.}
We consider various DNN architectures to comprehensively demonstrate the effectiveness of our attack and to evaluate how the architecture affects the attack performance.
Specifically, we consider VGG-16 (VGG16) \cite{simonyan2014very}, ResNet-18/50 (RN18/50) \cite{he2016deep}, WideResNet-50x4 (WRN50x4) \cite{wideresnet}, and DenseNet-161 (DN161) \cite{huang2017densely}.

\noindent \textbf{Training.}
We evaluate both vanilla training (\ie, minimizing the loss on clean inputs) and AT (\ie, minimizing the loss on adversarial examples) to demonstrate how AT makes it more easier for the privacy attack to succeed.
Specifically, in this work we consider Madry's AT \cite{Advt_madry} with $\ell_2$ norm bound, although our attack can work with other types of AT (\eg, TRADES \cite{zhang2019theoretically}, or with $\ell_{\infty}$ norm bound) as both attack steps only rely on AT's general properties.
We also vary the perturbation strength of the AT to see its effect on the attack.
We leverage the pre-trained models provided by \cite{salman2020adversarially} to conduct all the experiments.

\noindent \textbf{Attack process.}
We simulate an attack process by first creating a set of \textit{anchor} images.
Specifically, we randomly sample 5 images from each of the 1,000 categories from the validation set of ImageNet, resulting in a total of 5,000 anchors.
Then, for each anchor, we construct 5 random batches which include that anchor.
Finally, we play as an attacker who tries to reconstruct each anchor given the 5 groups of batch-averaged gradients corresponding to the 5 random batches.
Note, here we focus on the \textit{best-case} performance out of the 5 batches, which equivalently exposes the \textit{worst-case} scenario for the victim/defender.
We explicitly focus on the reconstruction of the anchor images because it allows easier analysis.
%\eg, we can analyze the vulnerability of different image categories to the attack.

\noindent \textbf{Evaluation metric.}
We measure the cosine similarity between the restored features and the ground-truth features to evaluate the performance of our first attack step, \ie, the quality of feature restoration.
When evaluating the second step, namely image reconstruction from features, we will both use LPIPS \cite{zhang2018unreasonable} as the quantitative metric and qualitatively demonstrate the performance by visualizing the reconstructed images.

\begin{figure}[!t]
  \centering
  \includegraphics[width=.99\linewidth]{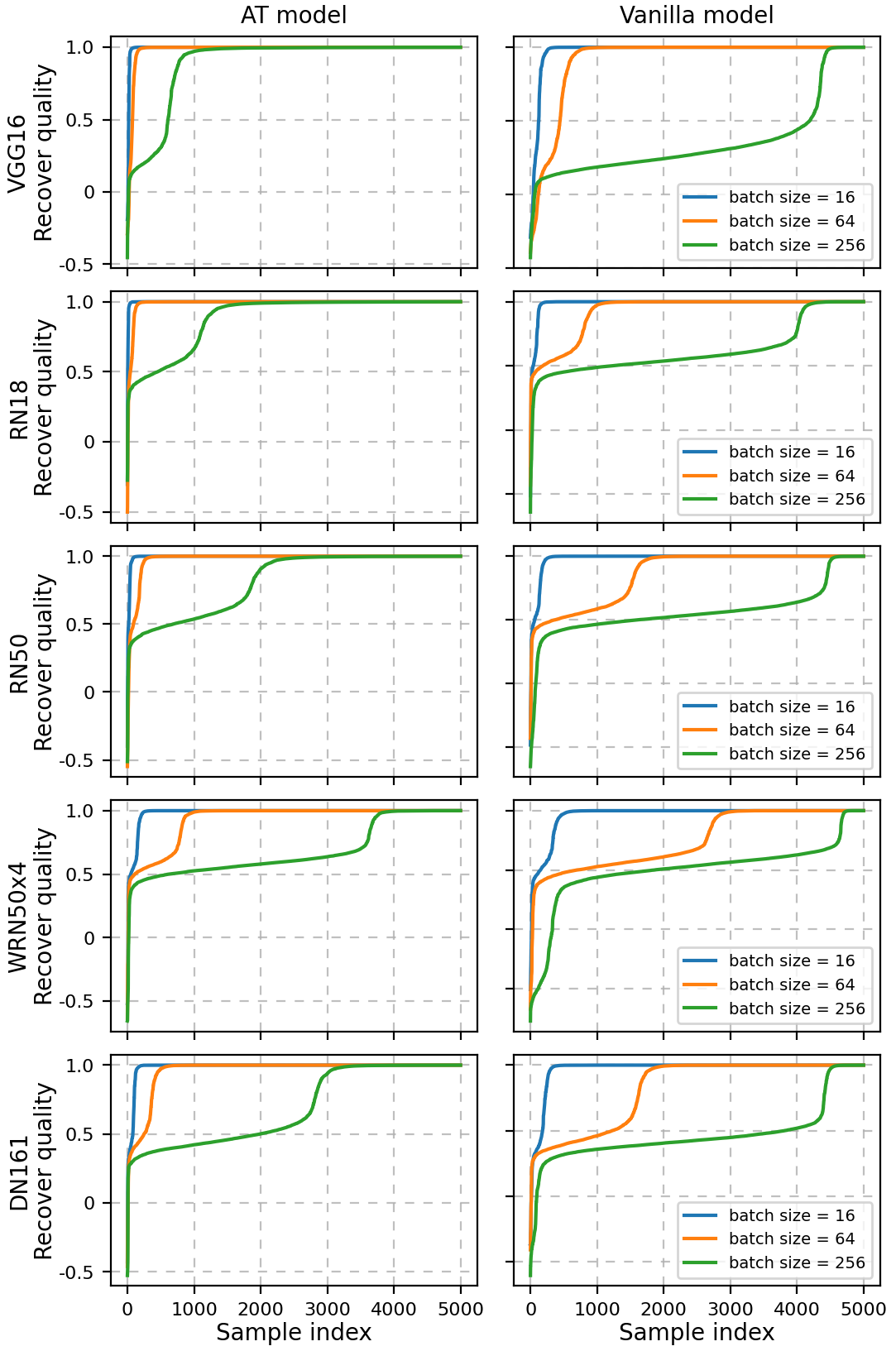}
  %\vspace{-1mm}
  \caption{Feature restoration quality (in terms of cosine similarity) of the 5,000 anchor images on various AT and vanilla models and under various batch size. The samples are sorted by the recover quality in each plot. The closer the curve is to the upper left corner, the better feature restoration is achieved.}
  \label{fig:feat_recov_result_1}
  \vspace{-4.5mm}
\end{figure}

\begin{figure}[!h]
  \centering
  \includegraphics[width=.95\linewidth]{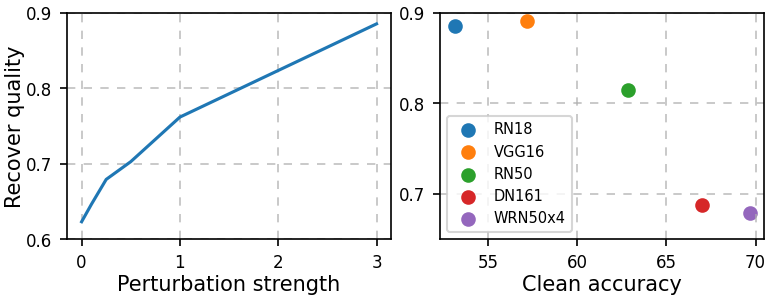}
  %\vspace{-1mm}
  \caption{\textbf{Left:} Feature restoration quality v.s. the perturbation strength $\varepsilon$ of AT. RN18 is used here. \textbf{Right:} Feature restoration quality v.s. the model architecture. $\varepsilon=3.0$ is used here.}
  \label{fig:feat_recov_result_2}
  \vspace{-6mm}
\end{figure}

\subsection{Feature restoration from gradients}

We first evaluate our first attack step, namely feature restoration, for which the results are shown in \cref{fig:feat_recov_result_1}. 
%Here we focus on the case where the perturbation strength $\varepsilon=3.0$, which is a standard configuration for AT on ImageNet \cite{salman2020adversarially}.
There are several key observations made from the results.

\noindent \textbf{AT models enable better restoration of the features.}
Comparing the first and second column of \cref{fig:feat_recov_result_1}, it is obvious that the feature restoration is much more successful on AT models than on vanilla models.
For example, with RN18 architecture and a batch size of 256, there are around 3,500 out of 5,000 samples whose features can be restored nearly perfectly (\ie, with $\sim 1.0$ cosine similarity) on the AT model, while less than 1,000 samples' features are restored that well on the vanilla model.
This observation aligns with our analysis in \cref{sec:4.2}.

In the left plot of \cref{fig:feat_recov_result_2}, we also observe that the feature restoration quality (averaged over 5,000 samples) becomes better as the AT's perturbation strength $\varepsilon$ increases, which further demonstrates that AT makes the model more vulnerable to the feature restoration attack step.

%For adversarial examples, the probability assigned to their ground-truth class $\bm{p}_{i,y_i}$ is much smaller than in the case of clean examples.
%Consequently, AT models can achieve a tighter approximation in \cref{eq:feat_restore} and thus lead to better feature restoration quality.

\noindent \textbf{Feature restoration works even under large batch size.}
As discussed earlier, the motivation of the feature restoration step is to decouple the fused information of each individual input from the batch-averaged gradients.
Indeed, we find this strategy effective even under a large batch size.
Specifically, under the batch size of 256, the attack applied to an AT model can perfectly restore the features of at least 1,000 samples (with WRN50x4) and up to 4,000 samples (with VGG16), out of the total 5,000 samples.
Not surprisingly, the restoration quality is even better when the batch size is smaller.

\noindent \textbf{Models with smaller capacity are more vulnerable.}
In the right plot of \cref{fig:feat_recov_result_2}, we visualize the feature restoration quality (averaged over 5,000 anchor samples) w.r.t. the test accuracy of various model architectures on the clean images of ImageNet.
The general trend is that the models with lower clean accuracy is more vulnerable to the feature restoration attack step.
We suspect that this is because the models with a smaller learning capacity (in terms of the clean accuracy) do not resist adversarial examples very well, and the probability $\bm{p}_{i,y_i}$ assigned to the ground-truth class of the adversarial input would be lower.
As a result, small-capacity models could lead to better feature restoration (see \cref{eq:feat_restore}).

\subsection{Image reconstruction from features}

\begin{figure}[!t]
  \centering
  \includegraphics[width=.99\linewidth]{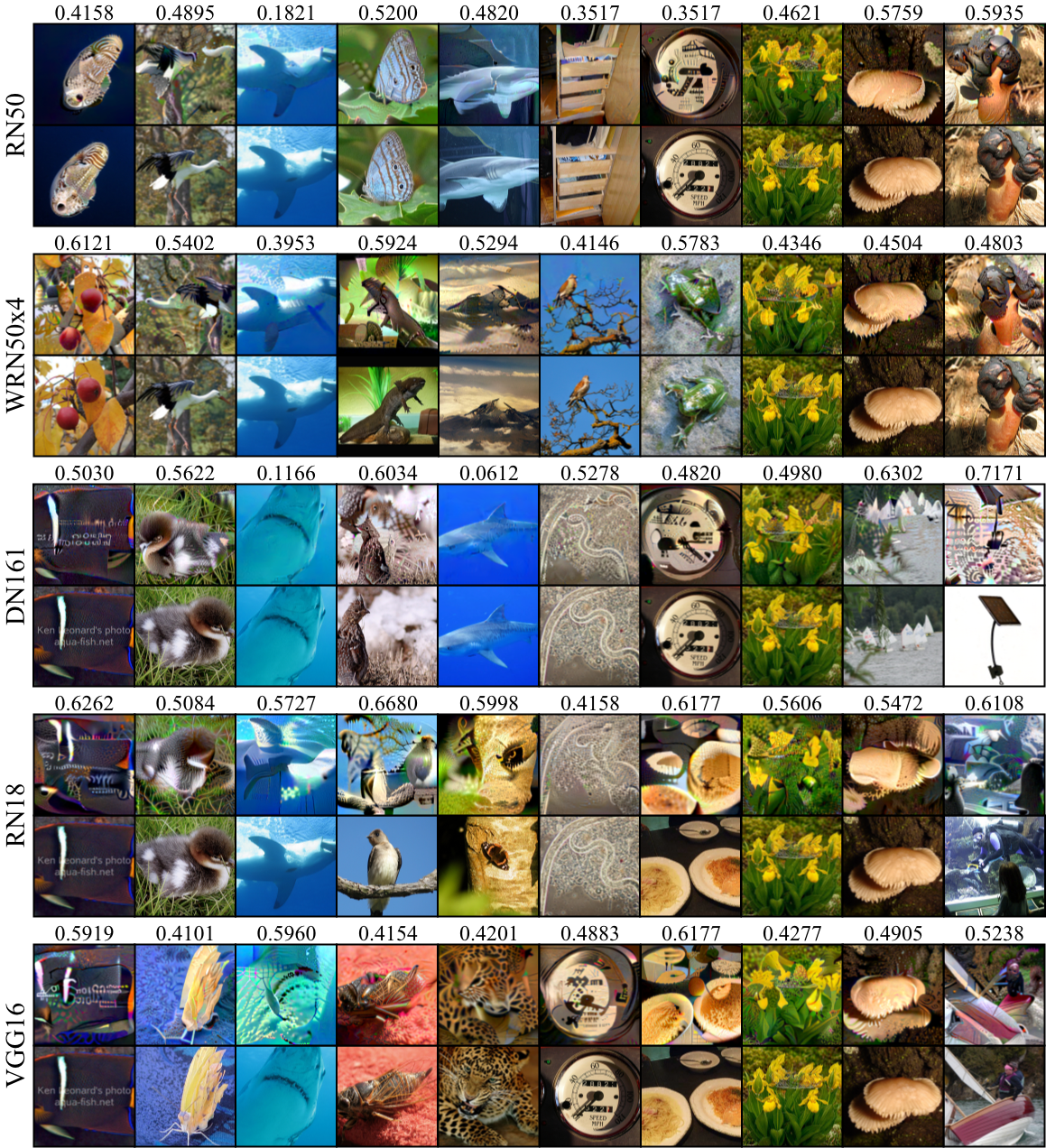}
  %\vspace{-1mm}
  \caption{Visualization of reconstructed images on AT models (first row in each pair) and ground-truth ones (second row). The numbers are the LPIPS score ($\downarrow$).}
  \label{fig:demo}
  %\vspace{-6.5mm}
\end{figure}

\begin{figure}[!t]
  \centering
  \includegraphics[width=.99\linewidth]{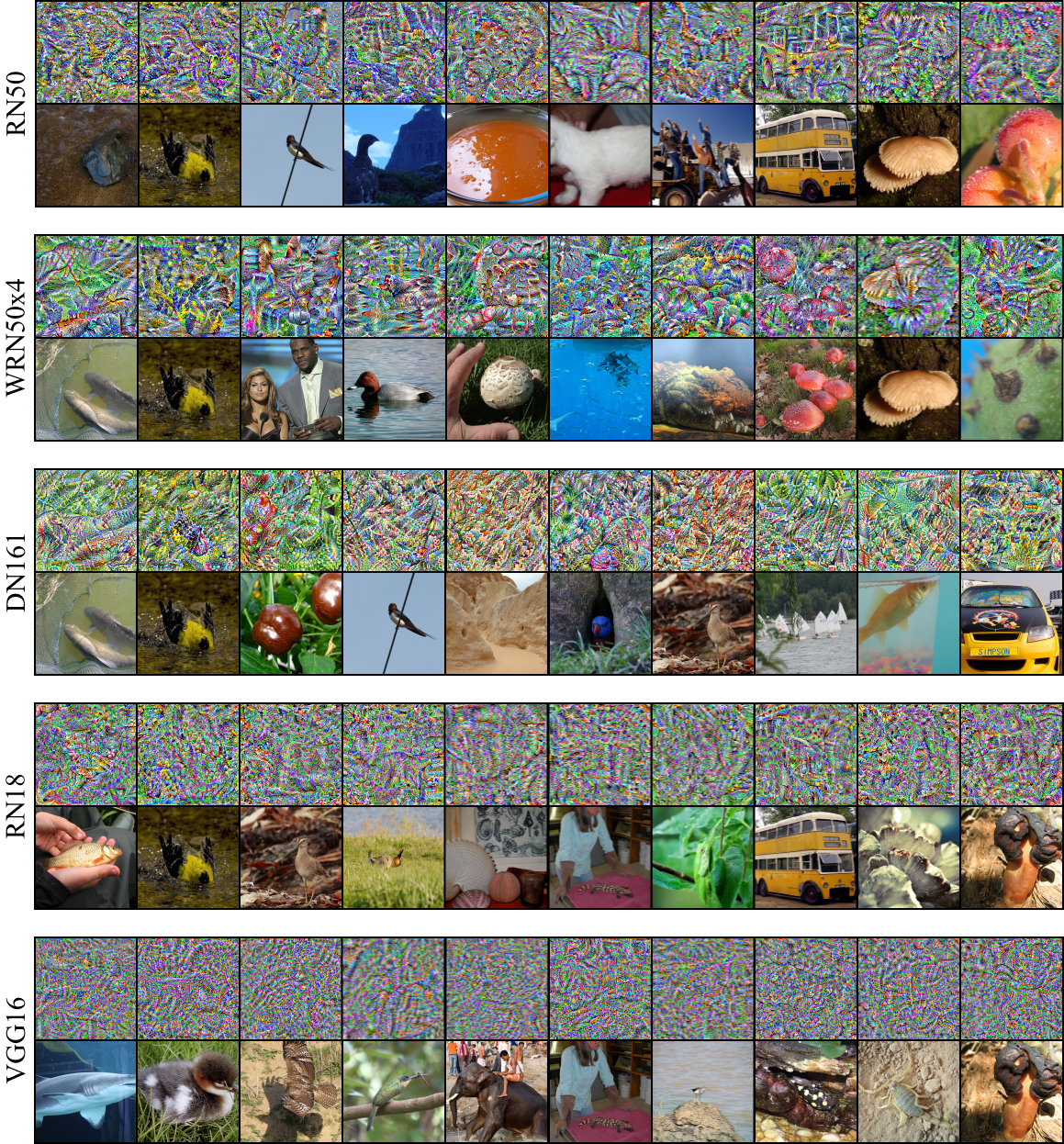}
  %\vspace{-1mm}
  \caption{Visualization of reconstructed images on vanilla models (first row in each pair) and ground-truth ones (second row).}
  \label{fig:demo_vanilla}
  %\vspace{-6.5mm}
\end{figure}

We next invert the restored features to recontruct input images according to \cref{eq:feat_2_image}.
Specifically, we use Adam optimizer with a learning rate of 0.1 and perform 5,000 steps of optimization.
The weight for the Total Variation loss $\mathcal{R}$ is $1e^{-6}$.
Image generation is often sensitive to the initialization, and we find that using multiple random starts (and picking the best result according to the loss) can improve the image quality.
Here we use 5 random starts.

We present the results obtained on AT models and vanilla models in \cref{fig:demo} and \cref{fig:demo_vanilla}, respectively.
The takeaway here is twofold.
First, AT models obviously enable much better reconstruction than vanilla models.
Many of the generated images from AT models clearly unveil the object identity, often with accurate restoration of the visual details (\eg, the branches in the bird image obtained on WRN50x4).
In comparison, vanilla models only yield absurd, hard-to-interpret patterns which hardly reveal useful information about the original images.
Second, we find that the models with larger learning capacity are more suitable for the feature inversion.
Recall that RN50, WRN50x4, and DN161 achieve higher clean accuracy than RN18 and VGG16 according to \cref{fig:feat_recov_result_2}.
In \cref{fig:demo}, the latter two models more often have distortions or unnatural patterns in their reconstructed images than the first three models.
We suspect that this is because the large-capacity models can learn more semantic-meaningful features \cite{engstrom2019adversarial} during AT, which is essential for recovering the semantics of the original input images.

Finally, we remark that with AT models, our attack can better compromise the clients' privacy than prior arts.
Specifically, our method achieves an averaged LPIPS score of $0.4424$ on RN50 under a batch size of 256, while the current state-of-the-art method \cite{yin2021see} obtained $0.4840$ under a batch size of 8 with the same model architecture.
In addition, to our knowledge we are the first to report success of reconstruction under the batch size of 256 among the privacy attacks towards FL systems.

\iffalse
- Focus on the case where batch size = 256. Again, for each of the 50,000 samples construct multiple random batches (if we do find recover accuracy is mainly a function of sample itself then maybe can just use one batch) and provide both qualitative and quantitative measure of recover quality.

For example, can make density plot of PSNR and LPIPS of all samples, and meanwhile visualizing a few samples with different PSNR/LPIPS values to give people a sense.

- The final image recovery definitely depends on the representation recovery. First empirically identify a threshold (e.g. >0.995?) beyond which the image recovery can be accurate. (Is this necessary?)

- Also suspect the recovery quality depends on the specific sample. Some sample might be naturally easier to reconstruct, while others are not.
\fi

%% file: sections/6_conclusion.tex
\section{Conclusion}

We develop a novel privacy attack against Adversarial Training models to break the privacy of Federated Learning systems.
Evaluation demonstrates that our attack can accurately reconstruct the clients' training images even when the batch size is large (up to 256).
Thus, the clients that perform AT in the pursuit of \textit{robustness} are in the same time putting their \textit{privacy} at risk.
By further exposing such robustness-privacy trade-off of AT models with a practical attack, we hope to motivate future studies to address this issue and develop some kind of privacy-aware AT.

%- We believe that the perceptually-aligned gradients/representation is the source of privacy leakage of AdvT models. This leakage is more severe when the attacker can obtain/recover the representation. Although we focus on FL settings, there are other possible vulnerable scenarios along the same line of thought. For example, when crowdsourcing data from users but an adversarially trained feature extractor is used to enable later robust training. In this case the attack is trivial since the representations of the data would be directly sent to the cloud and might be acquired by the attacker.

%- We hope our work can motivate future studies on the trade-off between adversarial robustness and privacy.

%% file: main.bbl
\begin{thebibliography}{10}\itemsep=-1pt

\bibitem{croce2020robustbench}
Francesco Croce, Maksym Andriushchenko, Vikash Sehwag, Edoardo Debenedetti,
  Nicolas Flammarion, Mung Chiang, Prateek Mittal, and Matthias Hein.
\newblock Robustbench: a standardized adversarial robustness benchmark.
\newblock {\em arXiv preprint arXiv:2010.09670}, 2020.

\bibitem{deng2009imagenet}
Jia Deng, Wei Dong, Richard Socher, Li-Jia Li, Kai Li, and Li Fei-Fei.
\newblock Imagenet: A large-scale hierarchical image database.
\newblock In {\em 2009 IEEE conference on computer vision and pattern
  recognition}, pages 248--255. Ieee, 2009.

\bibitem{engstrom2019adversarial}
Logan Engstrom, Andrew Ilyas, Shibani Santurkar, Dimitris Tsipras, Brandon
  Tran, and Aleksander Madry.
\newblock Adversarial robustness as a prior for learned representations.
\newblock {\em arXiv preprint arXiv:1906.00945}, 2019.

\bibitem{fredrikson2015model}
Matt Fredrikson, Somesh Jha, and Thomas Ristenpart.
\newblock Model inversion attacks that exploit confidence information and basic
  countermeasures.
\newblock In {\em Proceedings of the 22nd ACM SIGSAC conference on computer and
  communications security}, pages 1322--1333, 2015.

\bibitem{geiping_inverting_grad}
Jonas Geiping, Hartmut Bauermeister, Hannah Dr\"{o}ge, and Michael Moeller.
\newblock Inverting gradients - how easy is it to break privacy in federated
  learning?
\newblock In H. Larochelle, M. Ranzato, R. Hadsell, M.~F. Balcan, and H. Lin,
  editors, {\em Advances in Neural Information Processing Systems}, volume~33,
  pages 16937--16947. Curran Associates, Inc., 2020.

\bibitem{he2016deep}
Kaiming He, Xiangyu Zhang, Shaoqing Ren, and Jian Sun.
\newblock Deep residual learning for image recognition.
\newblock In {\em Proceedings of the IEEE conference on computer vision and
  pattern recognition}, pages 770--778, 2016.

\bibitem{huang2017densely}
Gao Huang, Zhuang Liu, Laurens Van Der~Maaten, and Kilian~Q Weinberger.
\newblock Densely connected convolutional networks.
\newblock In {\em Proceedings of the IEEE conference on computer vision and
  pattern recognition}, pages 4700--4708, 2017.

\bibitem{KeskarMNST17}
Nitish~Shirish Keskar, Dheevatsa Mudigere, Jorge Nocedal, Mikhail Smelyanskiy,
  and Ping Tak~Peter Tang.
\newblock On large-batch training for deep learning: Generalization gap and
  sharp minima.
\newblock In {\em 5th International Conference on Learning Representations,
  {ICLR} 2017, Toulon, France, April 24-26, 2017, Conference Track
  Proceedings}. OpenReview.net, 2017.

\bibitem{Advt_madry}
Aleksander Madry, Aleksandar Makelov, Ludwig Schmidt, Dimitris Tsipras, and
  Adrian Vladu.
\newblock Towards deep learning models resistant to adversarial attacks.
\newblock In {\em 6th International Conference on Learning Representations,
  {ICLR} 2018, Vancouver, BC, Canada, April 30 - May 3, 2018, Conference Track
  Proceedings}. OpenReview.net, 2018.

\bibitem{mahendran2015understanding}
Aravindh Mahendran and Andrea Vedaldi.
\newblock Understanding deep image representations by inverting them.
\newblock In {\em Proceedings of the IEEE conference on computer vision and
  pattern recognition}, pages 5188--5196, 2015.

\bibitem{FL}
Brendan McMahan, Eider Moore, Daniel Ramage, Seth Hampson, and Blaise~Aguera y
  Arcas.
\newblock Communication-efficient learning of deep networks from decentralized
  data.
\newblock In {\em Artificial intelligence and statistics}, pages 1273--1282.
  PMLR, 2017.

\bibitem{mejia2019robust}
Felipe~A Mejia, Paul Gamble, Zigfried Hampel-Arias, Michael Lomnitz, Nina
  Lopatina, Lucas Tindall, and Maria~Alejandra Barrios.
\newblock Robust or private? adversarial training makes models more vulnerable
  to privacy attacks.
\newblock {\em arXiv preprint arXiv:1906.06449}, 2019.

\bibitem{salman2020adversarially}
Hadi Salman, Andrew Ilyas, Logan Engstrom, Ashish Kapoor, and Aleksander Madry.
\newblock Do adversarially robust imagenet models transfer better?
\newblock {\em Advances in Neural Information Processing Systems},
  33:3533--3545, 2020.

\bibitem{shokri2017membership}
Reza Shokri, Marco Stronati, Congzheng Song, and Vitaly Shmatikov.
\newblock Membership inference attacks against machine learning models.
\newblock In {\em 2017 IEEE symposium on security and privacy (SP)}, pages
  3--18. IEEE, 2017.

\bibitem{simonyan2014very}
Karen Simonyan and Andrew Zisserman.
\newblock Very deep convolutional networks for large-scale image recognition.
\newblock {\em arXiv preprint arXiv:1409.1556}, 2014.

\bibitem{song2019privacy}
Liwei Song, Reza Shokri, and Prateek Mittal.
\newblock Privacy risks of securing machine learning models against adversarial
  examples.
\newblock In {\em Proceedings of the 2019 ACM SIGSAC Conference on Computer and
  Communications Security}, pages 241--257, 2019.

\bibitem{adv_example}
Christian Szegedy, Wojciech Zaremba, Ilya Sutskever, Joan Bruna, Dumitru Erhan,
  Ian~J. Goodfellow, and Rob Fergus.
\newblock Intriguing properties of neural networks.
\newblock In Yoshua Bengio and Yann LeCun, editors, {\em 2nd International
  Conference on Learning Representations, {ICLR} 2014, Banff, AB, Canada, April
  14-16, 2014, Conference Track Proceedings}, 2014.

\bibitem{tramer2020adaptive}
Florian Tramer, Nicholas Carlini, Wieland Brendel, and Aleksander Madry.
\newblock On adaptive attacks to adversarial example defenses.
\newblock {\em Advances in Neural Information Processing Systems},
  33:1633--1645, 2020.

\bibitem{yin2021see}
Hongxu Yin, Arun Mallya, Arash Vahdat, Jose~M Alvarez, Jan Kautz, and Pavlo
  Molchanov.
\newblock See through gradients: Image batch recovery via gradinversion.
\newblock In {\em Proceedings of the IEEE/CVF Conference on Computer Vision and
  Pattern Recognition}, pages 16337--16346, 2021.

\bibitem{wideresnet}
Sergey Zagoruyko and Nikos Komodakis.
\newblock Wide residual networks.
\newblock In Edwin R.~Hancock Richard C.~Wilson and William A.~P. Smith,
  editors, {\em Proceedings of the British Machine Vision Conference (BMVC)},
  pages 87.1--87.12. BMVA Press, September 2016.

\bibitem{zhang2019theoretically}
Hongyang Zhang, Yaodong Yu, Jiantao Jiao, Eric Xing, Laurent El~Ghaoui, and
  Michael Jordan.
\newblock Theoretically principled trade-off between robustness and accuracy.
\newblock In {\em International conference on machine learning}, pages
  7472--7482. PMLR, 2019.

\bibitem{zhang2018unreasonable}
Richard Zhang, Phillip Isola, Alexei~A Efros, Eli Shechtman, and Oliver Wang.
\newblock The unreasonable effectiveness of deep features as a perceptual
  metric.
\newblock In {\em Proceedings of the IEEE conference on computer vision and
  pattern recognition}, pages 586--595, 2018.

\bibitem{zhao2020idlg}
Bo Zhao, Konda~Reddy Mopuri, and Hakan Bilen.
\newblock idlg: Improved deep leakage from gradients.
\newblock {\em arXiv preprint arXiv:2001.02610}, 2020.

\bibitem{zhu2020deep}
Ligeng Zhu and Song Han.
\newblock Deep leakage from gradients.
\newblock In {\em Federated learning}, pages 17--31. Springer, 2020.

\end{thebibliography}
